\documentclass{article}

\usepackage{arxiv}
\usepackage[utf8]{inputenc} 
\usepackage[T1]{fontenc}    
\usepackage{hyperref}       
\usepackage{url}            
\usepackage{booktabs}       
\usepackage{amsfonts}       
\usepackage{nicefrac}       
\usepackage{microtype}      
\usepackage{lipsum}
\usepackage{gensymb}
\usepackage{amsmath}
\usepackage{booktabs}
\usepackage{amsfonts}
\usepackage{csquotes}
\usepackage{array}
\usepackage{subcaption}
\usepackage{graphicx} 
\usepackage{url}            

\title{Adaptable Deformable Convolutions for Semantic Segmentation of Fisheye Images in Autonomous Driving Systems}

\author{
  Cl\'{e}ment Playout \\
  Department of Informatics\\
  \'{E}cole Polytechnique de Montr\'{e}al\\
  Montréal, Canada \\
  \texttt{clement.playout@polymtl.ca} \\
   \And
 Ola Ahmad\\
  CortAIx Thales, \\
  Montr\'{e}al, Canada\\
  \texttt{ola.ahmad@ca.thalesgroup.com} \\
  \And
  Freddy Lecue \\
  CortAIx Thales \\
  Montr\'{e}al, Canada\\
  Inria \\
  Sophia Antipolis, France\\
  \texttt{freddy.lecue@inria.fr} \\
  \AND
  Farida Cheriet \\
  Department of Informatics\\
  \'{E}cole Polytechnique de Montr\'{e}al\\
  Montréal, Canada \\
  \texttt{farida.cheriet@polymtl.ca} \\
}
\DeclareMathOperator{\taninv}{arctan}

\begin{document}
	\maketitle
	
\begin{abstract}
	Advanced Driver-Assistance Systems rely heavily on perception tasks such as semantic segmentation where images are captured from large field of view (FoV) cameras. State-of-the-art works have made considerable progress toward applying Convolutional Neural Network (CNN) to standard (rectilinear) images. However, the large FoV cameras used in autonomous vehicles produce \enquote{fisheye} images characterized by strong geometric distortion. This work demonstrates that a CNN trained on standard images can be readily adapted to fisheye images, which is crucial in real-world applications where time-consuming real-time data transformation must be avoided. Our adaptation protocol mainly relies on modifying the support of the convolutions by using their deformable equivalents on top of pre-existing layers. We prove that tuning an optimal support only requires a limited amount of labeled fisheye images, as a small number of training samples is sufficient to significantly improve an existing model's performance on wide-angle images. Furthermore, we show that finetuning the weights of the network is not necessary to achieve high performance once the deformable components are learned. Finally, we provide an in-depth analysis of the effect of the deformable convolutions, bringing elements of discussion on the behavior of CNN models.
\end{abstract}
\section{Introduction}
Convolutional Neural Networks have established state of the art performances on numerous vision-based tasks (object detection, instance and semantic segmentation) and datasets and therefore become a standard for many application. In particular, these models are major assets for Advanced Driver-Assistance Systems (ADAS) and autonomous vehicles, which require models that are highly efficient while allowing a good understanding of their operation. ADAS primarly rely on a precise identification of the environment surrounding the vehicle. In order to capture this environment, different types of sensors are used, among which fisheye cameras. These are built with specific lenses to achieve extremely wide field-of-view (FoV), reaching angle far superior to regular lenses. As a comparison, the latter are considered wide angle when covering angles from 64\degree to 84\degree, whereas the former can reach values up to 270\degree (but usually between 100\degree  to 180\degree). However, such high FoVs come at the expense of the rectilinear property provided by regular lenses (straight features in the scene remain straight in the image). Fisheye lenses capture curvilinear images, often roughly described as deformed by barrel distortion (the magnification decreases with the distance from the optical axis -usually the center of the image). Due to their nature, fisheye images are usually harder to analyse with conventional recognition models, requiring these to be tuned and/or retrained. In the context of deep learning, this usually comes to changing the training data, but this raises the question of the number of labelled fisheye images needed and whether the model can still benefits from existing rectilinear datasets. In this work, we demonstrate the impact of using these datasets in order to improve fisheye segmentation by analysing different ways to adapt an existing \enquote{rectilinear} CNN model to fisheye images. We consider different possibilities, from training a model from scratch with distorted fisheye-like images to simply slightly modifying the support of the convolution using trainable convolutions offsets (known as deformable convolutions).
As fisheye data is not a largely open resources in the Machine Learning community, we investigate the minimal number of samples needed to fulfill model adaptation.
For the sake of simplicity, the focus of this paper is on semantic segmentation but we hypothetize that most of the technics aftermentioned would generalize to object detection.

\noindent\textbf{Contribution 1:} We propose a novel and simple learning mechanism to adapt standard CNN models on fisheye images using the concept of deformable convolutions. To capture non-linear transformations, we demonstrate through this mechanism the possibility of learning the spatial support of convolution filters independently from their weights while maintaining almost similar performance. This insight could potentially motivate a rethinking of the data augmentation process in deep learning applications.

\noindent \textbf{Contribution 2:} We demonstrate that adapting a CNN model to the non-linear spatial distortion induced by ultra wide-angle geometry can be achieved with only a few training images. In this way, we can alleviate the lack of available training datasets of real fisheye images for different perception tasks. 

\noindent \textbf{Contribution 3:} As advanced driver-assistance systems commonly make use of narrow as well as large FoV cameras to perform different autonomous tasks, we propose a flexible semantic segmentation model that can be deployed with both narrow and large FoVs.  

\section{Related work}
From a computer vision researcher's perspective, the field of fisheye images is relatively recent, in particular when considering tasks involving object recognition (detection or semantic segmentation). The initial studies on the subject were mainly focused on calibration (finding the intrinsic parameters of the camera), by building a geometrical or analytical model. \cite{scaramuzzaFlexibleTechniqueAccurate2006a} proposed a parametric model following a polynomial form to describe the projection function. A fourth-order polynomial was proved to be an accurate model. This model is still regularly used, including to synthesize fisheye distortion in rectilinear images. Calibration is a potential first step toward image rectification, whose goal is to remove the distortion such that straight-line features appear as straight lines in the image.
Recently, \cite{yinFishEyeRecNetMultiContextCollaborative2018b} proposed to use a CNN to directly rectify fisheye image, by training the network to predict the distortion parameters. The authors demonstrated that using a semantic context as an input to the distortion parameters predictor significantly improves the system's performances. 
This suggests that rectification could be used as a preprocessing step before object recognition using an existing conventional model. Nonetheless, there is not much work in the literature proposing to combine both processes. 
As pointed out by \cite{yogamaniWoodScapeMultiTaskMultiCamera2019}, this can be explained by the fact that undistortion causes major issues: a typical fisheye with FoV$>90\degree$ can not be mapped onto a rectilinear image, leading to a reduction of the FoV in the rectified image. Moreover, the re-warping operation leads to a non-uniform sampling accross the resulting image, creating blurry areas. Therefore, current research is shifting to focusing on model adaptation rather than undistorting images. To our knowledge, \cite{fuchengdengObjectDetectionPanoramic2017} are the first to use a standard CNN architecture for objects detection in fisheye-like images. In their case, the images are treated as normal ones and used to finetune a network trained on rectilinear images. The adaptation therefore consists in changing the training data rather than on the model itself. This approach is comparable to what \cite{saezRealTimeSemanticSegmentation2019, yeUniversalSemanticSegmentation2020} proposed, which is to rely on synthetic fisheye generation to extend the training distribution. By sampling different distortion parameters, this types of approaches replicate the effect of data augmentation. 
We argue that adapting a model pretrained on rectilinear images should not require a full retraining and propose the idea that only a limited amount of new data if needed to implicitly learn the parameters of the image distortion and thereby adapt the model's semantic prediction. This intuition is motivated by \cite{lopezDeepSingleImage2019}, who showed that it is possible to predict extrinsinc and intrisinc camera distortion parameters from a single image. By extension, our work aims to explore the number of training samples needed to adapt a semantic segmentation model to interpret distorted images. For this task, using a single distorted image for finetuning would likely result in biasing the internal statistics of the model and eventually to strong model overfitting. Therefore, instead of modifiying the weights and biases of the model itself, we propose to change the way convolutions are done using deformable convolutions, as originally introduced by \cite{daiDeformableConvolutionalNetworks2017}. Deformable components should theoretically be able to capture the distortion parameters of the image, while the regular convolutions should be able to extract meaningfull semantic features. This idea has also been explored by \cite{dengRestrictedDeformableConvolutionBased2019}, who transformed deformable convolutions into their restricted equivalent and tested them on fisheye images. Their work also analyze the placement of the deformable layers within the network to achieve optimal performance.
Our approach differs from theirs as we only adapt the deformable part of the convolutions, rather than training a complete model. We also study in depth the effect of the positioning of deformable convolutions within the existing network, the effect of batch-normalization and explore few-shots training approaches.

\section{Methodology}
\paragraph{Problem Statement:} Given a set $S=\{X, Y\}$, where $X=\{x^{(i)}\}_{i=1}^{N}$ is a set of rectilinear images,  $Y=\{y^{(i)}\}_{i=1}^{N}$ the set of their associated 2D semantic segmentation groundtruth and $N$ the number of samples in the dataset, given a semantic segmentation model $\phi$, its prediction $\phi(x)=\tilde{y}$ and given a parametrized conversion function $\rho$ that associates a rectilinear image to its fisheye equivalent, this works targets to find the adaptation function $\Psi$ that optimizes:
\begin{equation}
\label{eqn:problemStatement}
\begin{split}
&\Psi(\phi) = \hat{\phi} \\
&\min_{\Psi}{||\hat{\phi}(\rho(x)) - \rho(\tilde{y})||}
\end{split}
\end{equation}

Note that the definition of $\rho$ is voluntarily vague, since $\rho$ can represent a synthetic projection function just as well as a change in the actual camera lens.
In the real world, $\rho$ is most likely unknown (and unused). However, we will assume that we know its form in the rest of the paper as $\rho$ is needed in absence of an existing labelled fisheye dataset. 
To summarize, we want the adapted model to predict from a wide angle image an output as close as possible to $\rho(\tilde{y})$, which is the distortion of the prediction obtained from a rectilinear image with the original model. \\
The rest of this section will describe each of the functions in equation \ref{eqn:problemStatement}, \textit{i.e.} the segmentation model $\phi$, the conversion function $\rho$ and finally the components of the adaptation function $\Psi$.

\subsection{Baseline for semantic segmentation}  Semantic segmention of rectilinear images has been thoroughly studied and refined to a point where many efficient CNN architectures are now easily deployable. The choice of a particular model is mainly based on a trade-off between performance requirements and computational resources available. We favour the former criterion and choose the DeepLabV3+ model as our baseline network. DeepLabV3+ is an architecture introduced by \cite{chenEncoderDecoderAtrousSeparable2018a}, extending their previous work on large-scale networks using atrous convolutions. The architecture is composed of three mains components. It starts with an encoder network that reduces the spatial resolution of the input while increasing its depth. The encoder is used to extract features from the input image. The lowest-level features are fed to the second component, the Atrous Spatial Pyramid Pooling (ASPP). It consists of several parallel convolution layers using different dilations rates, working overall as a multi-scale convolutional layer. The last component of the architecture is a decoder module, that expands the features from the encoder and the ASPP back up to the input dimensions. Inspired by the skipped connections proposed by \cite{ronnebergerUNetConvolutionalNetworks2015}, the decoder concatenates low- and mid-level features from the encoder and outputs a segmentation map. We experimented with two different models as a backbone for the encoder, the \textit{resnet101} proposed by \cite{xieAggregatedResidualTransformations2017}
and the \textit{Aligned-Xception} introduced by \cite{cholletXceptionDeepLearning2017}. We did not observe any improvements with the latter and thus kept the former. The \textit{resnet101} was pretrained on a subset of the COCO train2017 dataset provided with the Torchvision library\footnote{ 
}.
\subsection{Synthesizing fisheye images} The lack of existing ultra wide-angle datasets has motivated many researchers to generate approximations from rectilinear images. One of the two following approaches is usually chosen:
\begin{itemize}
	\item Simulating a fisheye-like distortion on rectilinear images.
	\item Rendering images from a 3D scene using a virtual fisheye cameras.
\end{itemize}

The first option, while being easy to setup, suffers from the drawbacks related to the grid sampling as mentioned in the previous section. 
The second option is significantly more time-consuming and requires knowledge about 3D graphics tools. On the other hand, it has the advantage of generating more realistic images in terms of their distortion and their FoV. Nonetheless, the 3D rendered images are far from being as detailed as real world images. For the sake of completness and reproducibility, we have experimented with both approaches. 

\begin{figure}
	\centering
	\begin{subfigure}{.33\textwidth}
		\includegraphics[width=.9\linewidth]{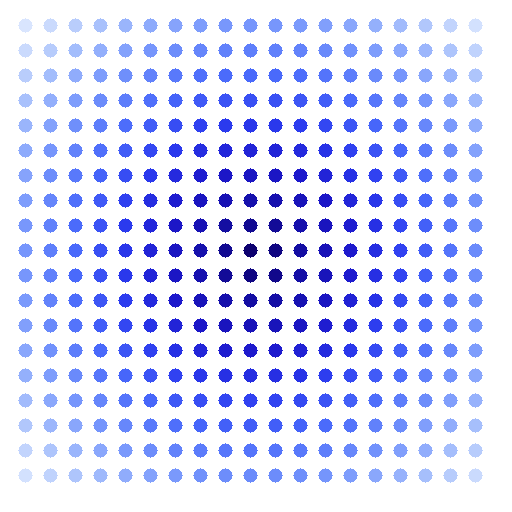}
		\caption{Original}
		\label{fig:sfig1}
	\end{subfigure}%
	\begin{subfigure}{.33\textwidth}
		\includegraphics[width=.9\linewidth]{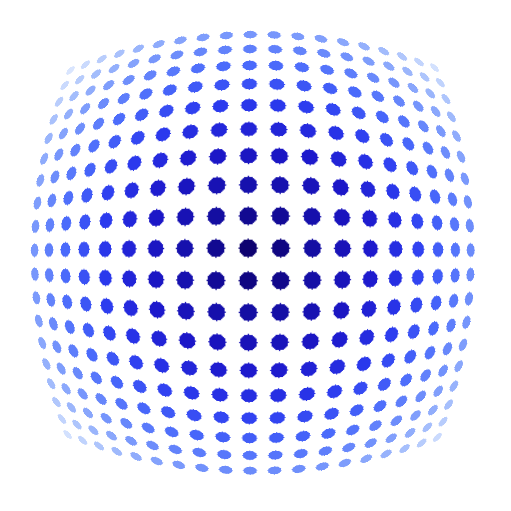}
		\caption{f=125}
		\label{fig:sfig2}
	\end{subfigure}
	\begin{subfigure}{.33\textwidth}
		\includegraphics[width=.9\linewidth]{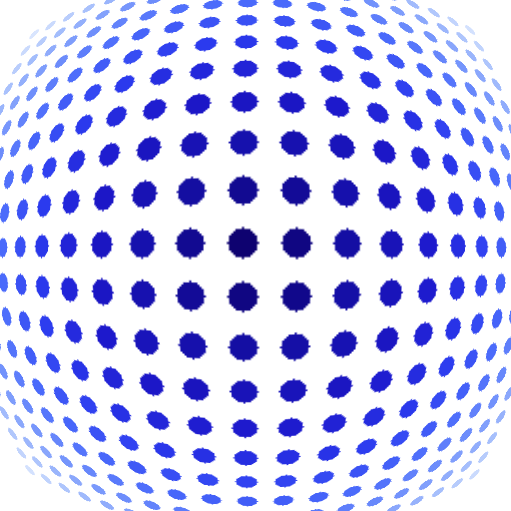}
		\caption{f=75}
		\label{fig:sfig2}
	\end{subfigure}
	\caption{Example of distortion using a parametric polynomial to synthesize fisheye-like images.}
	\label{fig:syntheticFisheye}
\end{figure}

\noindent \textbf{Simulation} To simulate fisheye distortion on rectilinear images, we rely on the same model as the one used in the open-source library OpenCV\footnote{\url{https://docs.opencv.org/master/db/d58/group__calib3d__fisheye.html}}. Noting $(x, y)$ a couple of normalized coordinates in the rectilinear image, the distortion functions maps them to normalized fisheye coordinates $(x', y')$ using the following equations:
\begin{equation}
\begin{split}
&r^2 = x^2 + y^2 \\
&\theta = \taninv(r)\\
&\theta_d = \theta\cdot(1+k_1\theta^2+k_2\theta^4+k_3\theta^6+k_4\theta^8) \\
&x'=f\cdot(\theta_d/r)\cdot x +x'_0\\
&y'=f\cdot(\theta_d/r)\cdot y +y'_0
\end{split}
\end{equation}
The parameters $\{f, k_i\}_{i=1}^{4}$ are tunable and $(x'_0, y'_0)$ can be adjusted to change the distortion center. We limit ourself experimentally to variations of $f$, which corresponds to a scale factor (as an approximation of a varying focal length), as depicted in Figure \ref{fig:syntheticFisheye}. Using this set of equations, we are able to apply the distortion on real images from the Cityscape dataset freely provided by \cite{cordtsCityscapesDatasetSemantic2016}.

\begin{figure}[t]
	\centering
	\begin{subfigure}{.49\textwidth}
		\includegraphics[width=\linewidth]{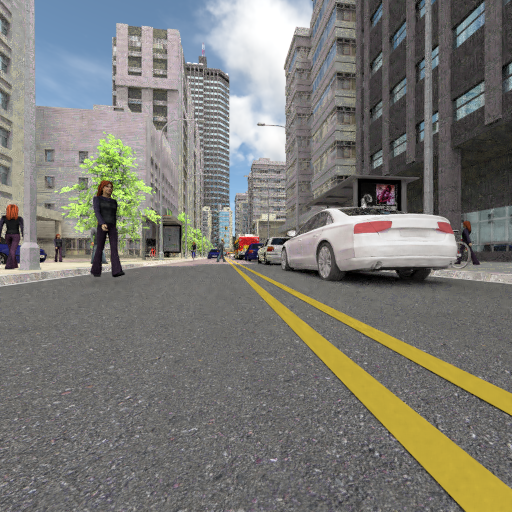}
		\caption{Rectilinear, FoV=80\degree}
		\label{fig:blenderRectilinear}
	\end{subfigure}
	\begin{subfigure}{.49\textwidth}
		\includegraphics[width=\linewidth]{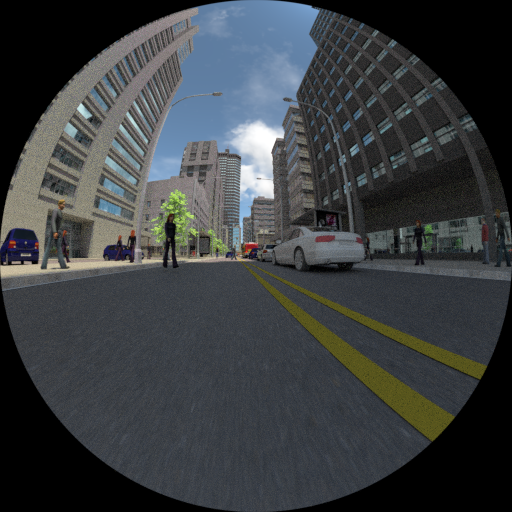}
		\caption{Fisheye, FoV=180\degree}
		\label{fig:blenderFisheye}
	\end{subfigure}
	\caption{3D renders from the same spatial camera position but two different lenses.}
	\label{fig:BlenderDataset}
\end{figure}

\noindent \textbf{Generation}
Our generated dataset is based on an extension of the 3D urban scene provided by \cite{zichaozhangBenefitLargeFieldofview2016}. Instead of the orignal depth-maps provided, we configure the render engine to output semantic maps and rendered images. Moreover, we enrich the scene by adding different 3D assets obtained from a free assets-provider\footnote{\url{https://www.blendswap.com/} All assets downloaded are free-to-use for non-commercial purposes.}. We thereby added the following objects to the existing scene: cars (6 different models), pedestrians (3 models), bikes (2 models), cyclists, bus and bus station. The limited variability of these scene elements might limit the usefulness of this dataset for real-world applications, but it can still provide important insights into fisheye segmentation. In total, the generated semantic maps are composed of 11 differents classes. Images are rendered at the resolution of $512\times512$. Two renders are done, the first one with a rectilinear camera (FoV=80\degree) and the second one with a fisheye camera (FoV=180\degree). As the same scene, with the same contents is represented for both, our comparative study is only focused on the effect of the fisheye distortion all else things being equal. Figure \ref{fig:BlenderDataset} shows the type of results we obtain. We refer to this dataset as \textit{BlenDataset}.

\subsection{Deformable convolutions}
\begin{figure}
	\centering
	\includegraphics[width=0.5\linewidth]{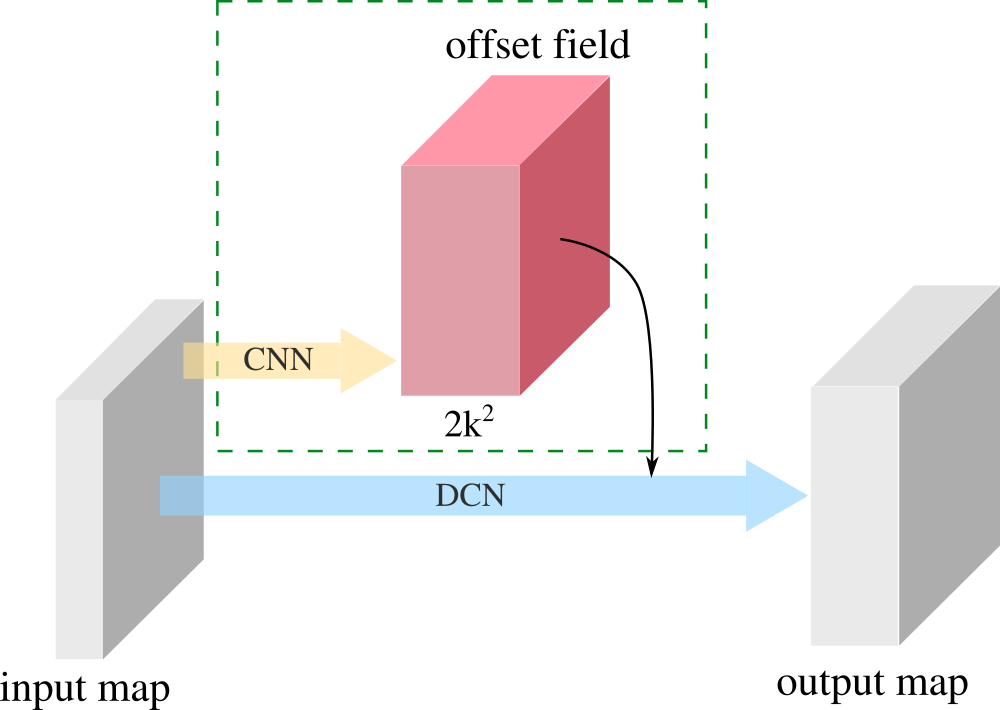}
	\caption{Deformable convolution (DCN) inserts themselves within a regular convolution layer with a $k \times k$ kernel. Our adaptation process only requires to train the prediction of the offsets (in light orange).}
	\label{fig:DCN_brick}
\end{figure}

Deformable convolutions (DCN) were introduced by \cite{daiDeformableConvolutionalNetworks2017} to extend the regular grid sampling locations used in convolutions with 2D offsets. DCN learn the offsets at every location from the feature maps of a previous layer, leading to dense representations that captures free deformation form of spatial kernels that apply in the current convolution layer. The offset simply refers to the displacement vector ($\mathbf{\delta p_i}$) of a point $\mathbf{p_i},  i\in \{1,...,n\times m\}$ taken on the $n\times m$ spatial grid of the kernel. 
Given a location $p$, where a standard convolution kernel applies, the output feature value of the deformable convolution layer at $\mathbf{p}$ becomes:
\begin{equation}
\label{eqn:DCN}
\mathbf{y}(\mathbf{p}) = \sum_{p_i \in \mathcal{R}} w(\mathbf{p_i})\cdot \mathbf{x}(\mathbf{p}+\mathbf{p_i}+\mathbf{\delta_{p_i}})
\end{equation}
where $\mathcal{R}$ is the grid support of the convolution.
Because of the fractional nature of the offsets values, \cite{daiDeformableConvolutionalNetworks2017}. used bilinear interpolation to apply Equation \ref{eqn:DCN}. Originally, the weights of the kernel were learned simultaneously with the offsets. The DCN was applied on rectilinear images to enhance convolutions in CNNs. In this work, we leverage the DCN's capabilities to take into account non-linear transformations brought by fisheye geometric distortion (as shown in Figure \ref{fig:syntheticFisheye}) and suggest an adaptable approach for fisheye image recognition tasks. To do so, we propose to transfer the weights of a base model (trained on rectilinear images) to the task of fisheye segmentation by converting regular convolutions to DCN and to only train the offsets layers that precede the main convolution layer (as shown in \ref{fig:DCN_brick}). The resulting model is called \textit{adaptable deformable convolution} as it adapts an existing convolution to the extrinsic deformations of the grid while preserving the intrinsic properties of the objects. Following equation \ref{eqn:problemStatement}, the objective is to find the optimal adaptation parameters of $\Psi$ that minimizes the error between the predicted output and groundtruth. The offsets are implicitly learned by backpropagating this error while the parameters of the base convolution layers remain fixed.

\section{Experiments}
\begin{figure*}[t]
	\centering
	\begin{subfigure}{.33\linewidth}
		\includegraphics[width=\linewidth]{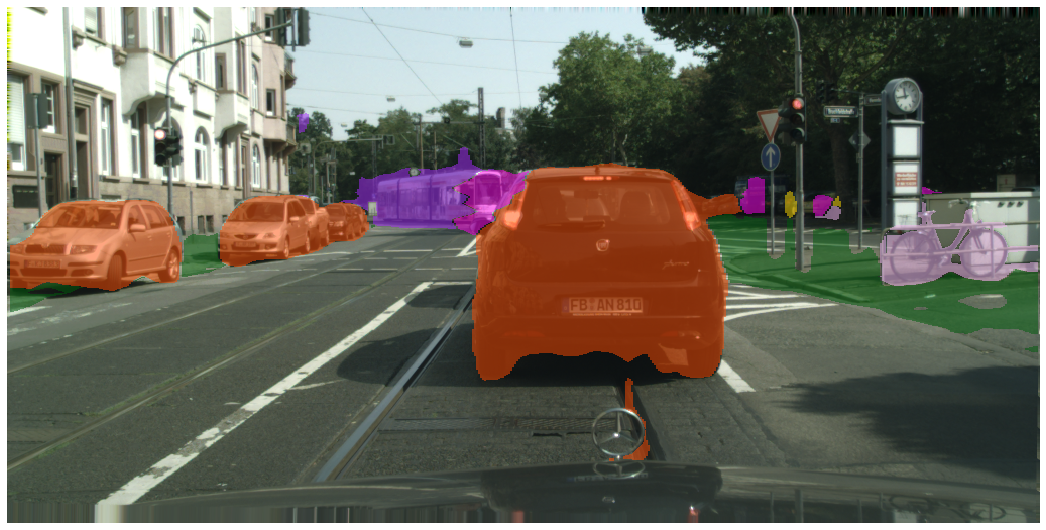}
		\caption{\textit{rect-DL3}}
	\end{subfigure}
	\begin{subfigure}{.33\linewidth}
		\includegraphics[width=\linewidth]{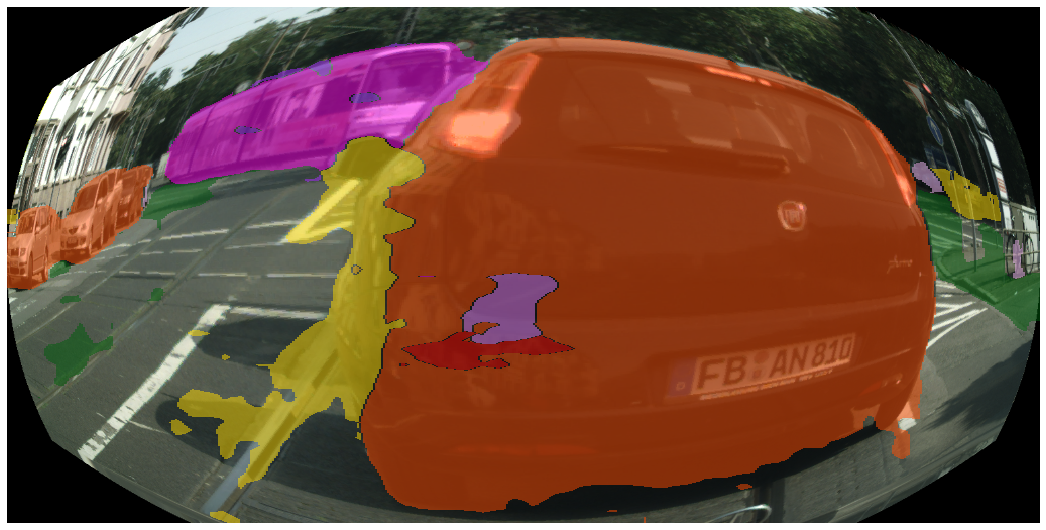}
		\caption{\textit{fish-DL3}}
	\end{subfigure}
	\begin{subfigure}{.33\linewidth}
		\includegraphics[width=\linewidth]{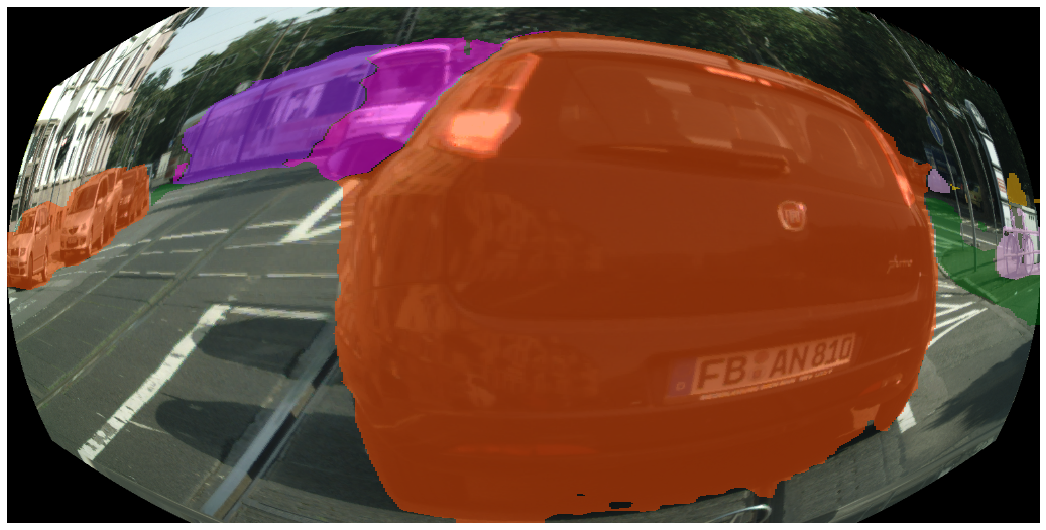}
		\caption{\textit{adpt-DL3}}
	\end{subfigure}
	\caption{Comparison of the prediction obtained from the same image with three different models. For the sake of readability, only 5 classes are shwon. The second and third image are distorted with $f=125$. \textit{rect-DL3} has been trained on rectilinear images, \textit{fish-DL3} has been trained from scratch on distorted images and \textit{adpt-DL3} is the adapted version of \textit{rect-DL3} (where convolutions are replaced by DCN but weights are kept equals)}
	\label{fig:finalQuali}
\end{figure*}

Two datasets were used independantly for our experiments, meaning that we trained different models using the same protocol for each dataset.

\noindent \textbf{BlenDataset:} This dataset is composed of 4000 pairs of rectilinear and fisheye synthetic images, as well as their corresponding groundtruth semantic segmentations. The dataset corresponds to a single video sequence taken in the 3D virtual scene. The first 3000 images were used for training purposes and the remaining 1000 images for testing. This was done in order to avoid having consecutive (and therefore highly correlated) frames in the two distinct sets. The training set was furthermore randomly split into two subsets (training and validation), following a 0.8/0.2 ratio.

\noindent \textbf{Cityscape:}
The Cityscape dataset comprises 5000 images divided into training, validation and test sets (2975, 500 and 1525 images respectively). Groundtruth maps are not publicly available for the test set, therefore we only used the first two sets. The validation set was used as the test set and the original training set was split in two (0.9/0.1 ratio) for training and validation purposes.\\
For all experiments, the images were kept at their original resolution ($1024\times 2048$), and random patches of size $512 \times 1024$ were extracted from them.

\noindent \textbf{Rectilinear segmentation training:}
We trained the DeepLabV3+ architecture on 4 GPUs, using synchronized batch-norm as proposed by \cite{pengMegDetLargeMiniBatch2018}, mixed-precision and a learning rate of 0.005 for the encoder and 0.05 for the decoder. Both learning rate were updated during training using the \enquote{poly} schedule policy introduced by \cite{liuParseNetLookingWider2015}. Weights were updated using the Adam solver. As a loss function, the weighted cross-entropy was used to alleviate the issue of class imbalance. We also observed that data augmentation (random scaling, rotation and horizontal flipping) helped to improve the model's performance. Training ran for 100 epochs, with a batch size of 8. We refer to this model as \textit{rect-DL3} (for \textit{\enquote{rectilinear DeepLabV3+}}).

\noindent \textbf{Adaptative training: }
In order to adapt the model, we have added deformable convolutions on top of the trained rectilinear model.
We use the efficient implementation from the \textit{MMDetection} toolbox provided by \cite{mmdetection}. 
Following the principle of an ablation study, we tested different configurations in order to understand the mechanisms underlying the different network components and demonstrate their respective effects. For each configuration, the adaptation was trained during 25 epochs on fisheye images (generated or simulated). To limit the scope of this paper, we used a fixed distortion level, parameterized by $f=125$.
We found that the prediction of DCN offsets was unstable when using high learning rates. Consequently, we reduced the rates to 0.01 for the decoder and 0.001 for the encoder. All others training parameters were kept identical as in the rectilinear training procedure described in the previous paragraph.

\noindent \textbf{Evaluation procedure}
As an evaluation metric, we have adopted the mean Intersection-over-Union (mIoU) proposed by \cite{cordtsCityscapesDatasetSemantic2016}. The IoU is computed per class for the whole test set and then averaged accross classes. For both datasets, a \enquote{void} class, corresponding to unsegmented objects and/or borders of the image, was included and corresponding pixels were discarded from the metric computations.
\subsection{Adaptation experiments}

\noindent \textbf{Why adapt a model?}
The need for model adaptation stems from the difficulty regular rectilinear models have in segmenting wide-angle images. This can be illustrated quite simply by directly testing \textit{rect-DL3} (without adaptation) on different fisheye distortions. The performances are reported in table \ref{tab:rectDeepLab3+}.
\begin{table}[h]
	\centering
	\begin{tabular*}{0.9\linewidth}{l@{\extracolsep{\fill}}cccc} 
		\toprule 
		\multicolumn{5}{c}{BlenDataset} \\
		\midrule
		& \multicolumn{2}{c}{Rectilinear} &  \multicolumn{2}{c}{Fisheye} \\
		\textit{rect-DL3} & \multicolumn{2}{c}{0.619} &  \multicolumn{2}{c}{0.517} \\
		\midrule
		\multicolumn{5}{c}{Cityscape} \\
		\midrule
		& Rectilinear &  $f=150$ & $f=125$ & $f=75$ \\
		\textit{rect-DL3} & 0.747 & 0.448 & 0.420 & 0.232 \\
		\bottomrule
	\end{tabular*}
	\caption{Performances (mIoU) obtained with two rectilinear models \textit{rect-DL3} (one per dataset) tested under different configurations. On Cityscape, fisheye effect is simulated and a lower value of f indicates a stronger distortion.} \label{tab:rectDeepLab3+}
\end{table}
As expected, they quickly deteriorate in comparison to the model's performance on rectilinear images. The performance degradation is correlated with the strength of the distortion.

\noindent \textbf{An Upper Limit to the Adaptation's Efficiency}
As the adaptation phase is only aimed at tackling the distortion problem, we can hypothesize an upper limit on the adapted model's performance. Given a metric function $\Omega$ (for example the mIoU), a rectilinear groundtruth map $y$, a prediction $\tilde{y}$ from a rectilinear input with \textit{rect-DL3} and a distortion function $\rho$, the best performance we can expect with the prediction $y_{adpt}$ from the adapted model on the corresponding distorted input is:
\begin{equation}
\Omega(y_{adpt}, \rho(y)) \leq \Omega(\rho(\tilde{y}), \rho(y))=\Lambda_\Omega
\end{equation}
In other words, for a distorted input, the best prediction possible corresponds to the distortion of the prediction obtained from the equivalent non-distorted input. This upper limit is denoted $\Lambda_\Omega$. With $f=125$ (the distortion level used in the following experiments), we obtain $\Lambda_{mIoU}=0.676$ with \textit{rect-DL3}. As expected, none of our adapted models exceeded this limit, but the adaptation process proved its efficiency by approaching the upper limit very closely (as shown in Table \ref{tab:Final} below). 

\noindent \textbf{Adapting with Batch Normalization}
Following the observations of \cite{liRevisitingBatchNormalization2016}, we noticed that finetuning the Batch Normalization (BN) layers during the adaptation phase helps to reach better performances than only learning the offset predictions. Hence, we investigated the effect of each component for the adaptation process; the results of this comparison are reported in Table \ref{tab:BN-effect}.

\begin{table}[h]
	\centering
	\begin{tabular*}{0.9\linewidth}{c@{\extracolsep{\fill}}cccc} \toprule 
		\multicolumn{5}{c}{Cityscape, $f =125$} \\
		\midrule
		\textit{rect-DL3} &  +BN & +DCN & +DCN+BN & $\Lambda_{mIoU}$\\
		0.420 & 0.527 &  0.531 & 0.643 & 0.676 \\
		\bottomrule
	\end{tabular*}
	\caption{Comparison of the adapation performances (mIoU) by finetuning batch-normalization (BN), training deformable offsets (DCN) and both.} \label{tab:BN-effect}
\end{table}
his experiment demonstrates that DCNs and BN are complementary options to adapt a given model. Nonetheless, it must be noted that tuning the BN on fisheye images might degrade the performance of the \textit{rect-DL3} model on rectilinear images, whereas the offsets of the DCNs' offsets can very easily be turned off, restoring the adapted model back to its original state. In autonomous vehicles that rely on both wide-angle and regular lenses, this ensures a very simple way to use the same model for both imaging modalities. 	

\noindent \textbf{What to Adapt?}
Similarly to \cite{daiDeformableConvolutionalNetworks2017} and \cite{dengRestrictedDeformableConvolution2019}, we studied which part of the original model needed to be adapted (\textit{i.e} adding DCN and finetuning the batch-norm) to obtain optimal performances. Rather than a layer-by-layer approach, we restricted the experiments to the two main blocks of the \textit{rect-DL3} model, the encoder and the decoder. Results are reported in Table \ref{tab:WhatToAdapt}.

\begin{table}[h]
	\centering
	\begin{tabular*}{0.9\linewidth}{c@{\extracolsep{\fill}}c@{\extracolsep{\fill}}c} \toprule 
		\multicolumn{3}{c}{Cityscape $f=125$} \\
		\midrule
		decoder only & encoder only & encoder+decoder\\
		0.414 & 0.639 & \textbf{0.643} \\
		\bottomrule
	\end{tabular*}
	\caption{Comparison of the effects of adapting different components of the models. For each case, the adaptation consists in adding deformable convolutions and tuning the batch-normalization.} \label{tab:WhatToAdapt}
\end{table}
As we can see, adapting the encoder and the decoder jointly seems to offer very little improvement over adapting only the encoder. This opens up interesting leads for future experiments, as a given encoder can be finetuned for different tasks (for example, ours could be tuned for object detection). In light of this, we believe that our adaptation method is likely to be suited for others tasks than semantic segmentation, meaning that the weights of the offset predictions could be re-used with no further tuning in an encoder trained for a different task such as object detection. The verification of this hypothesis is left for future work. 

\begin{figure}
	\centering
	\includegraphics[width=0.75\linewidth]{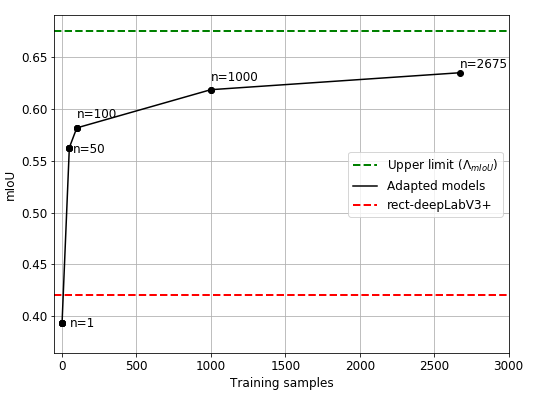}
	\caption{Performances (mIoU) with respect to the number of training samples.}
	\label{fig:FewShotLearning}
\end{figure}

\noindent \textbf{Few-shots adaptative training}
One of our central ideas of this work is that learning to adapt a model from distorted samples is significantly easier than learning semantic segmentation on them from scratch. We explored this notion by turning our adaptation protocol into a few-shots learning problem. Subsets of different sizes (1, 50, 100, 1000) from the initial training set were sampled and we evaluated the performance of the adapted model on the same test set. Training and testing were done on Cityscape ($f=125$). Results of this experiment are shown in Figure \ref{fig:FewShotLearning}. This graph shows that a single image training won't allow any improvement compared to the non-adapted model (mIoU dropping from 0.420 to 0.390). Nonetheless, even a relatively small number of samples (n=50 in Figure \ref{fig:FewShotLearning}) brings the  model's performance relatively close to the level reached when using the full training set (composed of 2675 images).

\noindent \textbf{Adaptation versus retraining}
Given the variability of existing fisheye simulation procedure as well as the heavy computational requirement to retrain different semantic models, we do not compare our adapted models directly with the state-of-the art. Nonetheless, we evaluate how it compared with a DeepLabV3+ trained on fisheye images with variable $f$ as a form of data augmentation as suggested by \cite{saezRealTimeSemanticSegmentation2019, yeUniversalSemanticSegmentation2020}. This model is referred to as \textit{fish-DL3}, to be distinguished from the adapted model, referred to as \textit{adpt-DL3}. Results are shown in Table \ref{tab:Final}. In addition, a qualitative comparison between the different models is shown in Figure \ref{fig:finalQuali}.
\begin{table}[h]
	\centering
	\begin{tabular*}{0.9\linewidth}{c @{\extracolsep{\fill}}ccc} 
		\toprule 
		\multicolumn{4}{c}{BlenDataset, fisheye} \\
		\midrule
		\textit{rect-DL3}& \textit{fish-DL3} & \textit{adpt-DL3} & 
		\\
		0.517 &0.588&0.615 & \\
		\midrule
		\multicolumn{4}{ c }{Cityscape, $f=125$} \\
		\midrule
		\textit{rect-DL3}& \textit{fish-DL3} & \textit{adpt-DL3} & $\Lambda_{mIoU}$ \\
		0.420 &0.514& 0.643 & 0.676 \\
		\bottomrule
	\end{tabular*}
	\caption{Comparison of performance (mIoU) between adapted model and retrained model.} \label{tab:Final}
\end{table}

\noindent \textbf{Discussion} The main experiments in this study served to demonstrate the effectiveness of the proposed adaptable deformable convolutions on fisheye images for semantic segmentation tasks. By learning only the weights of offset layers, the DCN-based model was able to adapt faster to non-linear spatial distortion and capture more accurate feature representations than finetuning or retraining a standard CNN. The proposed approach should remain valid for any choice of $f$, but experiments on dynamic values of $f$ should be conducted to explore the model's capacity to generalize to different types of fisheye cameras used in ADAS systems and autonomous vehicles. The offsets are learned by backpropagating the cross-entropy loss function. An unanswered question emerges: can we learn the offsets in a self-supervised way, thereby removing the need for explicit data annotation? Given the fact that real-world fisheye images are often difficult to annotate because of their distortion, self-supervised learning could save significant time and costs in adapting CNNs to vision-based tasks on ultra wide-angle images. This work contributes to the goal of self-supervised learning by providing the concept of an upper performance limit on model adaptation to fisheye images. We keep this research area to our future work. We see two limitations of our current work: (1) lack of direct comparisons with related work and (2) lack of validations on real fisheye images. The rationale behind these limitations is the availability of the data. To the best of our knowledge, existing methods (even very few) have used fisheye simulations with a different experimental setup and train on larger dataset to get high performance. In the absence of a unified dataset or at least a standardized distortion approach, fair comparisons are hard to make. To prevent this in the future, we plan on releasing soon our code and trained models.
\section{Conclusion and Future Work}   
Deformable convolutions have been shown to be a significant improvement over regular convolutions in many tasks. This work focuses on proving that they can be effectively used on top of an existing CNN without modifying its pre-trained weights. This opens up interesting applications for systems relying on multiple imaging modalities, as a single model can be reliably adapted to the different tasks by means of marginal modification rather than full retraining.
Moreover, we demonstrate that training the deformable components can be done independently from the rest of the model (even if finetuning the batch normalization is advised) and that it does not require a large number of samples and alleviates the need to build large datasets of labeled fisheye images. These observations open different avenues worth exploring in future work. In particular, since autonomous vehicles require real-time object detection, we plan to investigate whether our adaptation protocol can be applied to existing object detection models. 

\paragraph{Acknowledgment} This work was supported by NSERC (Natural Sciences and Engineering Research Council of Canada). The authors gratefully acknowledge Philippe Debann\'e for revising this manuscript.

\bibliographystyle{unsrt}  
\bibliography{references}

\end{document}